\documentclass{svproc}
\usepackage{graphicx}
\usepackage{subfiles}
\usepackage{cmap}
\usepackage{svg}

\usepackage{amsmath} 
\usepackage{hyperref} 
\usepackage{comment} 
\usepackage{cite} 

\usepackage{url}

\makeatletter
\newcommand{\printfnsymbol}[1]{%
  \textsuperscript{\@fnsymbol{#1}}%
}
\makeatother

\begin{document}
\mainmatter              
\title{Hybrid Graph Embedding Techniques in Estimated Time of Arrival Task
}
\titlerunning{Hybrid Graph Embedding Techniques in Estimated Time of Arrival Task}  
%
\author{Vadim Porvatov\inst{1,2}\printfnsymbol{1} \and
Natalia Semenova\inst{1}\thanks{equal contribution} \and
Andrey Chertok\inst{1,3}}
\authorrunning{V. Porvatov et al.} 
%
%
\institute{Sberbank, Moscow 117997, Russia, \and
National University of Science and Technology ``MISIS'', Moscow 119991, Russia, \and
Artificial Intelligence Research Institute (AIRI),\\
\email{eighonet@gmail.com}\\
\email{semenova.bnl@gmail.com}\\
\email{achertok@sberbank.ru}\\
\url{}}

\maketitle              
\begin{abstract}
Recently, deep learning has achieved promising results in the calculation of Estimated Time of Arrival (ETA), which is considered as predicting the travel time from the start point to a certain place along a given path. ETA plays an essential role in intelligent taxi services or automotive navigation systems.
A common practice is to use embedding vectors to represent the elements of a road network, such as road segments and crossroads. Road elements have their own attributes like length, presence of crosswalks, lanes number, etc. However, many links in the road network are traversed by too few floating cars even in large ride-hailing platforms and affected by the wide range of temporal events. As the primary goal of the research, we explore the generalization ability of different spatial embedding strategies and propose a two-stage approach to deal with such problems. 



\end{abstract}
\section{Introduction}

The modern state of traffic induces a remarkable number of forecasting challenges in a variety of related areas. According to the industrial needs, a relevant computation of the estimated time of vehicle arrival can be considered as one of the most actual problems in the logistics domain. In particular, intelligent traffic management systems \cite{6531823} require significant accuracy in case of arrival time estimation. Besides such an application, computation of ETA also appears as a common issue in the commercial areas which are strongly dependent on optimal routing. The explicit examples of such services are taxi \cite{distribution_cat}, railway \cite{eta_trains}, vessels \cite{PARK2021100012} and aircraft transportation \cite{eta_planes}. 

\begin{figure}[t]
\includegraphics[width=\textwidth]{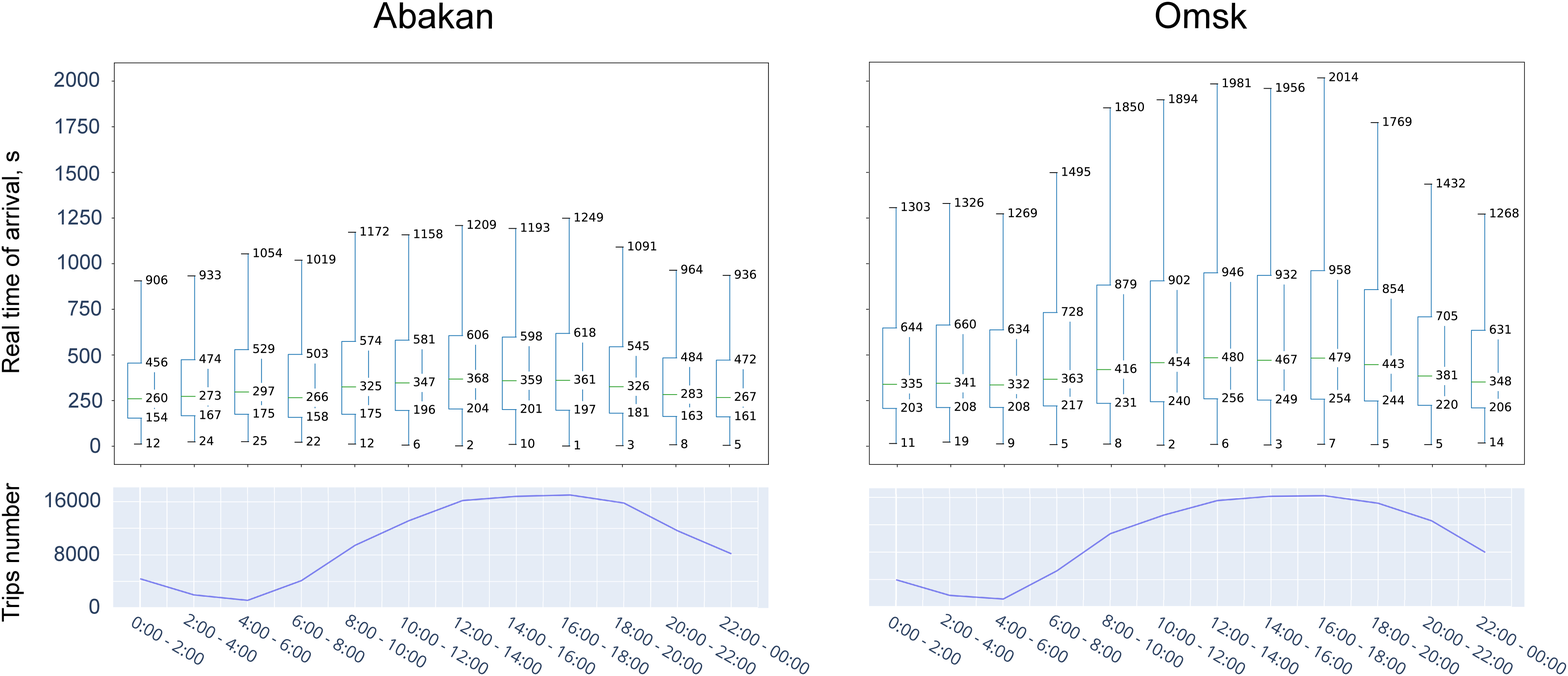}
\caption{Demonstration of temporal traffic dynamics: cumulative frequencies of car activity and distribution of trips duration for Abakan and Omsk in the two hours interval.} \label{usage_freqs}
\end{figure}

Accurate prediction of ETA for cars is a complex task requiring the relevant processing of heterogeneous data. It is frequently represented as time series and graph structure with feature vectors associated with its nodes and/or edges. In comparison with other vehicles, computation of ETA for cars is considerably influenced by the road network topology, nonlinear traffic dynamics, unexpected temporal events, and unstable weather conditions, Figure \ref{usage_freqs}. The stochastic nature of the introduced problem requires an implementation of a powerful domain-specific regression model with a high generalization ability. 

Machine learning proved its outstanding efficiency in a wide range of regression tasks. However, not every model can be efficiently applied to the ETA forecasting due to the mentioned constraints of available data. Previously performed attempts of a simple model implementation (e. g., linear regressions and gradient boosting) were reported as inefficient \cite{lr, WDR}, while the more sophisticated approaches allowed to achieve more optimistic results \cite{sun_2020_metric_learning_eta}. Thus, in order to obtain a better performance, we assume the necessity of applying graph neural networks \cite{gnn} as a part of the presented pipeline. 

According to the extensive growth of graph machine learning in recent years, many promising architectures \cite{kipf2017semisupervised, Perozzi_2014} emerged and soon were applied in a wide range of graph-related studies \cite{sweet_physics, protein_chem}. These models quickly became useful in terms of feature extraction in downstream tasks. Applied to the underlying graph structure of a city road network, such algorithms have the potential to dramatically increase the expressiveness of regression models and therefore should be explored. 

In the present paper, we propose and compare different architectures of the hybrid graph neural network for ETA prediction. Our main contributions are the following:

\begin{itemize}
    \item We introduce and publish the first to our best knowledge dataset\footnote[1]{to receive an access to data you need to send a request to semenova.bnl@gmail.com} with intermediate trip points. This dataset is relevant for consistent ETA prediction task and future usage as a benchmark. We provide common information about trips and city road network as well as road structural properties, marking, and weather conditions (other features are described in Section 3 in detail). Additionally, the route data includes auxiliary information which can be used both for evaluation of the ETA and independent prediction of real traveled distance as a separate problem.     
    \item Absence of methodological review of subgraph embeddings in the domain of interest encourages us to overwhelm such a limitation. Instead of focusing on more general approaches which include both spatial and recurrent temporal aspects, we prefer to precisely explore the domain of spatial embeddings as an underdeveloped one at the present moment.  
    \item We conduct a comprehensive evaluation of our method on two real-world datasets which correspond to tangibly different cities. Obtained results of computational experiments motivate us to further develop our research in accordance with achieved significant performance improvements.
\end{itemize}

\section{Related work}
As it has been mentioned above, the ETA-related tasks are a fundamental part of logistic services. In overwhelming number of cases, they demand two properties from the predictive algorithms: computational efficiency and relevant accuracy. The first part of this challenge was unequivocally solved by simple learning models like gradient tree boosting, multi-layer perceptron, and linear regression. However, the quality of these models cannot be reported as sufficient even beyond the commercial logistics. 

Along with the simple learning models, deterministic algorithms were also developed in huge amount \cite{10.1145/2820783.2820836, 10.1145/2623330.2623656}. In the majority of cases they cannot be compared with learning models in terms of quality. However, some of them were inspiring enough to influence the future development of their concepts in a more sophisticated way. 

\begin{figure}
\includegraphics[width=\textwidth]{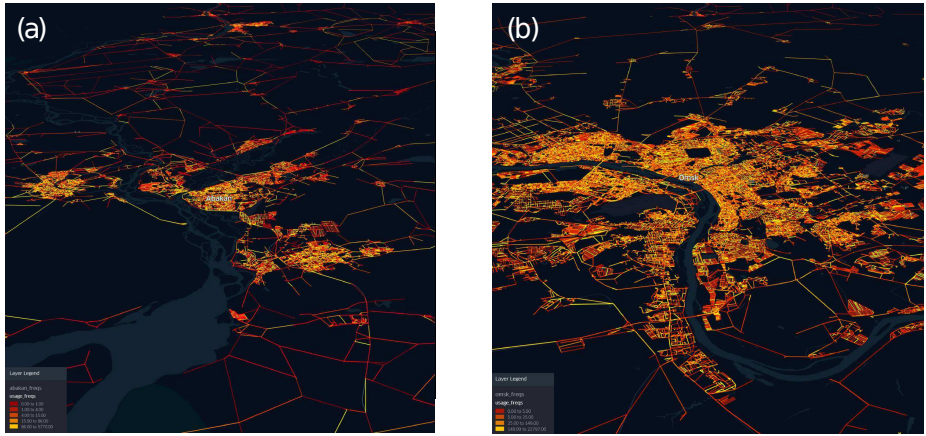}
\caption{Edges usage frequencies projected as a heatmap on the road networks of Abakan (a) and Omsk (b). The patterns of edges demand are clearly distinguishable as the topology of networks remains significantly different.} \label{projection}
\end{figure}

Limitations of mentioned approaches were partially overwhelmed in DeepTTE \cite{deeptte} and MURAT \cite{murat}. The first approach includes a recurrent neural network (RNN) which subsequently predicts the travel time along the trip. As many other recent methods, this algorithm is dependent on intermediate GPS coordinates. At the same time, the second method is closely related to the proposed architecture in the sense of graph embedding usage. In spite of the deep development of the temporal forecasting part, no more than one spatial embedding method was observed in any of this papers.

The most recent studies introduce new solutions with the potential to significantly increase the quality of ETA prediction. WDR \cite{WDR} is a wide-deep architecture that outperformed a lot of previously established approaches. Its further improvement and computational experiments led the same authors to the design of RNML-ETA architecture \cite{sun_2020_metric_learning_eta} which allows to achieve even better results. Simultaneously, another intriguing paper \cite{hstgcn} emerged as a prospective modification of ST-GCN methods family\cite{stgcn, astgcn, stsgcn}. All of these methods use datasets with intermediate points in contrary to the overwhelming majority of early papers. Following this positive trend, we continue studies in the same direction. 

\section{Data}
In the present work, we use two datasets related to the city networks of Abakan and Omsk. The cities have significantly different scales. Hence, their infrastructure pattern cannot be compared directly. Such a diversity allows us to check the generalization ability of the proposed architectures in a more explicit way. General properties of the dataset are established in Table 1 when the frequencies of road network segments usage are represented in Figure \ref{projection}(a, b). 

\begin{table}[b]
\caption{Description of the datasets in terms of common networks characteristics}
\begin{center}
\begin{tabular}{l@{\quad}@{\hspace{6em}}l@{\hspace{6em}}l}
\hline
\multicolumn{1}{l}{\rule{0pt}{12pt}Property}&
                   \multicolumn{1}{l}{Abakan}&
                   \multicolumn{1}{l}{Omsk}\\[2pt]
\hline\rule{-3pt}{12pt}
Nodes  &    65524 &   231688\\
Edges  &   340012 &    1149492\\
Total trips number  &   119986 &  120000\\
Trips coverage  &   0.535 &   0.392\\
Edges usage median  & 12  &   8\\[2pt]
\hline
\end{tabular}
\end{center}
\end{table}

Each dataset consists of both road networks and the routes associated with their edges. City networks contain an abundant number of meaningful features that can be translated to the predictive model in different ways. The route sample includes information about the start and destination point and a set of visited nodes during the ride. 

The trip data was collected in the period from December 1, 2020 up to December 31, 2020 by subsidiary companies of Sberbank. A comprehensive description of the proposed data is given in Table 2 for the city network and in Table 3 for car routes. 

\begin{table}
\begin{center}
\caption{Edge features of city network}
\begin{tabular}{l@{\quad}l@{\hspace{1.0em}}l}\hline
\multicolumn{1}{l}{\rule{0pt}{12pt}Feature}&
                   \multicolumn{1}{l}{Values}&
                   \multicolumn{1}{l}{Description}\\[2pt]
\hline\\[-8pt]\rule{-3pt}{12pt}
Road class & \begin{tabular}[c]{@{}l@{}}fake road, intra-quarter driveway, \\ dirt road, other city street, main \\ city  street, highway, intercity \\ road,  federal highway, cycle path, \\ walkway\end{tabular} & \begin{tabular}[c]{@{}l@{}}General road segments \\ categories\end{tabular} \\
Length & $\mathbf{Z_{+}}$ & \begin{tabular}[c]{@{}l@{}}Length of a road \\ segment in meters\end{tabular} \\
Width & $\mathbf{Z_{+}}$ & \begin{tabular}[c]{@{}l@{}}Width of a road \\ segment in meters\end{tabular} \\
Def  speed & \{3, 15, 20, 60, 90\} & \begin{tabular}[c]{@{}l@{}}Speed limit on a \\ road section in km/h\end{tabular} \\
Lanes & \{0, 1, 2, 3, 4, 5\} &  \begin{tabular}[c]{@{}l@{}}Number of lanes in \\ a road segment\end{tabular} \\
Barrier & \{0, 1\} & \begin{tabular}[c]{@{}l@{}}Defines the presence \\ of road barriers\end{tabular} \\
Payment flag & \{0, 1\} & \begin{tabular}[c]{@{}l@{}}Defines a road segment \\ as toll\end{tabular} \\
Turn restrictions & \{0, 1\} &   \begin{tabular}[c]{@{}l@{}}Defines an ability to \\ turn on a road section\end{tabular} \\
Pedo offset & \{0, 1\} & \begin{tabular}[c]{@{}l@{}}Defines the presence of \\ crosswalk offsets\end{tabular} \\
Bad road & \{0, 1\} &  \begin{tabular}[c]{@{}l@{}}Defines the condition \\ of a road segment\end{tabular} \\
Style & \begin{tabular}[c]{@{}l@{}}undefined, archway, crosswalk,\\ stairway, bridge, overground way,\\ invisible, normal, park path,\\ park footpath, subway, pedestrian \\ bridge, underground way, tunnel, \\ living zone, ford\end{tabular} & \begin{tabular}[c]{@{}l@{}}Additional road segments \\ categories\end{tabular}  \\[29pt]
\hline
\end{tabular}
\end{center}
\end{table}

\begin{table}
\begin{center}
\caption{Features of trip dataset}
\begin{tabular}{l@{\quad}@{\hspace{2.5em}}l@{\hspace{2.8em}}l}\hline
\multicolumn{1}{l}{\rule{0pt}{12pt}Feature}&
                   \multicolumn{1}{l}{Values}&
                   \multicolumn{1}{l}{Description}\\[2pt]
\hline\rule{-3pt}{12pt}
Nodes &$\{\hat{V} \subset V\}$ & \begin{tabular}[c]{@{}l@{}}Subset of nodes\end{tabular} \\
Dist to a & $\mathbf{Z_{+}}$ & \begin{tabular}[c]{@{}l@{}}Length of a segment between actual start \\ point and its projection on the first edge\end{tabular} \\
Dist to b & $\mathbf{Z_{+}}$ & \begin{tabular}[c]{@{}l@{}}Length of a segment between actual end \\ point and its projection on the last edge\end{tabular} \\
Start point part & $\mathbf{Z_{+}}$ & \begin{tabular}[c]{@{}l@{}}Part of the first edge where the trip \\ starts in meters\end{tabular} \\
Finish point part & $\mathbf{Z_{+}}$  &  \begin{tabular}[c]{@{}l@{}}Part of the last edge where the \\ trip ends in meters\end{tabular} \\
Start UTC & $\mathbf{Z_{+}}$ & \begin{tabular}[c]{@{}l@{}} Start time of the trip in UTC format\end{tabular} \\
Real time of arrival & $\mathbf{Z_{+}}$ & \begin{tabular}[c]{@{}l@{}}Trip duration in seconds\end{tabular} \\[2pt]
Real dist* & $\mathbf{Z_{+}}$ &   \begin{tabular}[c]{@{}l@{}} Actual traveled distance in meters \end{tabular} \\
Rebuild count* & $\mathbf{Z_{+}}$ & \begin{tabular}[c]{@{}l@{}}Number of route rebuilds that corresponds \\ to the destination change \end{tabular} \\[2pt]
\hline
\end{tabular}
\end{center}
\end{table}

According to the complexity of input data, it cannot be directly translated to a predictive model as an input. In order to correctly solve the desired task, it is recommended to filter the established dataset and perform feature engineering. Trips that have a rebuild count more than 1 should be optionally separated from the main volume of routes as well as anomaly short and long routes. Values of start (finish) point parts and dist to a(b) can be also added or subtracted from the total estimated length of the route in order to obtain a better spatial resolution of subgraph embeddings. 

\section{Methods}
The task can be mathematically formulated as a regression problem that extended by a special procedure of an automatic feature engineering. In order to handle this challenge, we generate vector representations of the road segments via GNNs, aggregate them to the trips embeddings and then apply a regression model which predicts ETA. 

Given a graph $G = (V, A, X)$ of the city road network, where $V = \{v_1, v_2, ... , v_n\}$ denotes the set of graph vertexes (road segments), $A$: $n\times n$ $\xrightarrow{} \{0, 1\}$ denotes the adjacency matrix (each edge encodes connectivity of the road segments), and $X$: $n \times m$ $\xrightarrow{}$ $\mathbf{R}$ is a matrix of node features.

The goal is to compute such a representation of each node $v_i \in V$ that can be effectively aggregated in accordance with structural properties of the route $s_j:= \{v_{j_1}, ..., v_{j_t}\}$, $s_j \in S$. There are two main aggregation strategies that potentially allow to construct a meaningful route subgraph embedding. The first one based on basic summation of all representations of the nodes that are included to the exact route
\begin{equation}
    z_{s_j} = \sum_{i=1}^{\#s_j} Z(v_{j_i}),
\end{equation}
where  Z($\cdot$) is the node embedding function.

Another approach related to initial graph extension by virtual nodes. This procedure induces a new graph $\hat{G}$($V'$, $A'$, $ X'$), where $V'$ = $\{v_1, ..., v_n, v_{n+1}, ...,$ $v_{n+\#S}\}$, $A'$:($n + \#S$)$\times$($n + \#S$) $\xrightarrow{}$ $\{0, 1\}$, $\forall v_i, v_n \in V$ adjacency matrix defined as $A'(v_i$, $v_j$) = $A$($v_i$, $v_j$). For the other edges, we propose the bijective function $f:V'$\textbackslash $V$ $\xrightarrow{}$ S that defines $\forall v'_{k} \in  V'$\textbackslash $V$ and $\forall v_l \in$ $f(v'_k)$ values in remaining part of the extended adjacency matrix as $A'(v_l$, $v'_k$) = 1. In agreement with this method,
\begin{equation}
    z_{s_j} = Z(f^{-1}(s_j)).
\end{equation}

For both strategies it is crucial to find the appropriate node embedding function $Z(\cdot)$ which has a significant impact on the relevance of the final route subgraph representations. We propose graph convolutional networks \cite{kipf2017semisupervised}, GAT \cite{gat}, and GraphSAGE \cite{graphsage} as the main candidates for nodes representation learning. The ideas behind these methods are quite similar as they all encode nodes to vectors of a fixed size via a repeated aggregation over a local neighborhood. However, while the GCN is based on mean aggregation, GraphSAGE pretends to be a more flexible and representative instrument due to its different aggregators and embedding concatenation stage. On the other hand, GAT adopts the mechanism of attention \cite{attention_first} firstly proposed in Natural Language Processing (NLP) to the needs of graph machine learning. To explicitly reveal the relevance of the mentioned approaches, in the following we briefly introduce the main aspects of each method.

\textbf{Graph Convolutional Network (GCN)}. For a given graph $G(V, A, X)$ this method defines an effective approach to network information aggregation. Single graph convolution layer is its atomic unit that can be represented as
\begin{equation}
H^{(l+1)}=\sigma\left(\tilde{D}^{-\frac{1}{2}} \tilde{A} \tilde{D}^{-\frac{1}{2}} H^{(l)} W^{(l)}\right),
\end{equation}
where $l+1$ is the current convolution layer number, $\sigma$ is an arbitrary nonlinear function (e. g., ReLU), $H^{(0)}$ = $X$, $\tilde{A}=A+I_{N}$, $\tilde{D}_{i i}=\sum_{j} \tilde{A}_{i j}$ and $W^l$ is the matrix of learning parameters.  

\textbf{GraphSAGE}. This algorithm mostly inherits the notation of convolutions from the GCN architecture, but instead of using full graph it directly computes convolution for each node $v$ in the iterative manner
\begin{equation}
h_{v}^{l+1} = \sigma\left(W^{l} \cdot \operatorname{CONCAT}\left(h_{v}^{l}, h_{N(v)}^{l+1}\right)\right),
\end{equation}
where $h_{N(v)}^{l+1}$ can be extracted by a few different aggregate functions for the set of neighbour nodes $N(v)$. 

\textbf{Graph Attention Network}. The last considered method is based on the attention mechanism which also avoids transductive GCN constraints and apply the iterative aggregation procedure 

\begin{equation}
h^{l+1}_i =\operatorname{CONCAT}_{k=1}^{K} \sigma\left( \; \sum_{j \in N(i)} \alpha_{i j}^{k} W^{k} h^{l}_{j}\right).
\end{equation}
The attention coefficient is computed as follows: 
\begin{equation}
\alpha_{i j}=\frac{\exp \left(\sigma\left(a^{T}\cdot\operatorname{CONCAT}(W h^{l}_{i}, W h^{l}_{j})\right)\right)}{\sum_{k \in N(i)} \exp \left(\sigma\left(a^{T}\cdot\operatorname{CONCAT}(W h^{l}_{i}, W h^{l}_{k})\right)\right)},
\end{equation}
where $a^T$ is a transposed vector of attention trainable parameters.

\begin{figure}
\includegraphics[width=\textwidth]{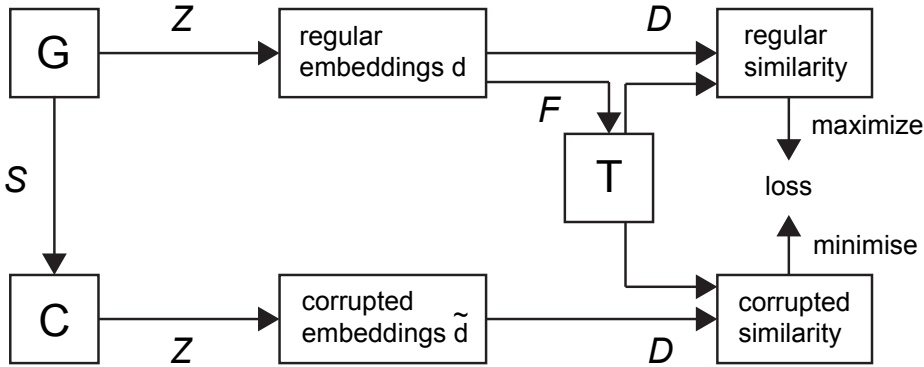}
\caption{Deep Graph Infomax corrupts feature vectors of the input graph G by function S (in the used realisation it shuffles features), constructs regular and corrupted node embeddings by applying $Z(\cdot)$, and finally estimates their similarity to the ground-truth vector T by the discriminator function D.} \label{dgi}
\end{figure}

In order to boost the expressiveness of these methods and convert supervised setups to unsupervised, we propose to embed them as a part of the Deep Graph InfoMax pipeline \cite{dgi}. This approach is based on minimizing of a two-component loss function
\begin{equation}
L=\frac{1}{N+M}\sum_{i=1}^{N} E_{G}\left[\log D(d_{i}, T)\right]+\sum_{j=1}^{M} E_{C}\left[\log \left(1-D(\tilde{d}_{j}, T)\right)\right]
\end{equation}
which aims to learn how to distinguish initial nodes representations $d$ and corrupted ones $\tilde{d}$, Figure \ref{dgi}.

Once embeddings of routes  $z_{s_j}$ are computed, each vector can be extended by additional information about the weather conditions and corresponding temporal categorical features. After these manipulations with route vectors $z_{s_j}$ they can be finally fed to the regression model.  

\section{Results}
In order to perform the training and evaluation of proposed architectures, we need to split the datasets into three samples. We trained our model on the first 100 000 trips, while the test and validation steps were performed on equal parts of the remaining datasets. 

Following the evaluation standards, we use a common set of metrics for the ETA prediction task: Mean Average Error (Eq. 8), Mean Average Percentage Error (Eq. 9), and Rooted Mean Square Error (Eq. 10).

\begin{equation}
\begin{aligned}
\mathrm{MAE} &=\frac{1}{N} \sum_{i=1}^{N}\left|y_{i}-y_{i}^{\prime}\right|,
\end{aligned}
\end{equation}
\begin{equation}
\begin{aligned}
\mathrm{MAPE} &=\frac{100}{N} \sum_{i=1}^{N}\left|\frac{y_{i}-y_{i}^{\prime}}{y_{i}}\right|, 
\end{aligned}
\end{equation}
\begin{equation}
\begin{aligned}
\mathrm{RMSE} &=\sqrt{\frac{1}{N} \sum_{i=1}^{N}\left(y_{i}-y_{i}^{\prime}\right)^{2}}.
\end{aligned}
\end{equation}

\subsection{Implementation details}

Computational experiments were provided with the use of StellarGraph\cite{StellarGraph} library. All models were trained on 2 GPU Tesla V100, the total training time of the pipeline for the best models is 9 hours. During the embedding construction process, we used three types of each observed architecture with the number of layers from 1 to 3 and the fixed output of size 128. Neural networks weights were trained by Adam optimizer\cite{Adam} due to its good convergence and stability. We use the static learning rate parameters $L_1$ = 0.001 for node embedding generation and $L_2$ = 0.0001 for regression.

\subsection{Experiments}

We performed series of computational experiments varying the strategy of subgraph embedding generation and the method of node representation extraction. As the final regression model, we leverage a multi-layer perceptron (MLP). 
For the purpose of Deep Graph InfoMax tests extension, we also compute the values of the metrics for regular unsupervised GraphSAGE and regression baseline to illustrate the general capabilities of different approaches. The final values of metrics for each configuration are shown in Table 4.

\begin{table}[]
\begin{center}
\label{results_table}
\caption{Evaluation results on test sample}
\begin{tabular}{l@{\quad}cccccc}
  \hline \rule{-4pt}{12pt}
  & \multicolumn{3}{c}{Abakan} & \multicolumn{3}{c}{Omsk}\\[2pt]
   \hline \rule{-4pt}{12pt}
                  & MAE & RMSE & MAPE & MAE & RMSE & MAPE\\
Baseline(MLP only) &  111.05 & 316.39 & 27.129 &  145.819 & 296.86  &  25.019\\
GraphSAGE + VN & 111.23 & 316.82 & 27.213 & 146.003 & 297.028 & 25.108 \\
GraphSAGE + Sum & 96.575 & 310.114  & 22.881  & \textbf{129.831} &\textbf{ 279.773} &\textbf{ 22.416}  \\
DGI(GCN) + Sum  & 97.927 & 310.628 & 23.506 & 141.017  & 289.32  & 24.335  \\
DGI(GAT) + Sum  & 101.808 & 313.01  & 25.737 & 133.262 & 283.22  &  23.175  \\
DGI(GS)  + Sum  & \textbf{95.819} & \textbf{309.627} & \textbf{22.622} & 130.296 & 280.058 & 22.593  \\
\hline
\end{tabular}
\end{center}
\end{table}

As it seen from the table, the best performance was achieved by the GraphSAGE setup with Deep Graph InfoMax in the case of Abakan. Meanwhile, common GraphSAGE also demonstrates promising embeddings quality (especially for Omsk) which is slightly different from its DGI modification. The error distributions of the best models for each dataset are shown in Figure 4.

\begin{figure}
\label{error}
\includegraphics[width=\textwidth]{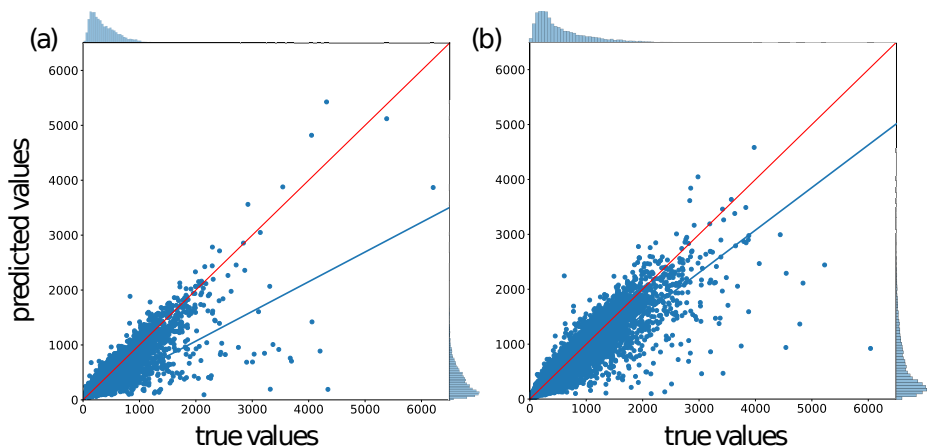}
\caption{Error distribution for the regression models trained on Abakan (a) and Omsk (b) datasets.} \label{confusion}
\end{figure}

Unfortunately, the test series of virtual nodes route embeddings turned down our pursuit to report any significant results. We conclude that the expressiveness of this method is limited in the area of interest, despite previous positive attempts of implementation in other tasks \cite{drugs}. However, such a result was partially foreordained by the studies which also explored subgraph embeddings \cite{subgnn}.

\section{Conclusion and Outlook}
In this work, we implemented and explored a pipeline that includes state-of-the-art algorithms of graph machine learning that emerged in recent years. We trained and tested our model on two consistent datasets which correspond to cities with different road topology types. Our results allow us to conclude that GraphSAGE-based models capture spatial patterns of city networks more substantially.    

Our own perspectives include future development and modification of more specific methods based on obtained results. As the primary goal of this research was to find the most efficient methods of subgraph embedding construction in the context of ETA problem, we intend to use this knowledge to construct a more complex spatial approach in the upcoming papers. In the spotlight of our research, we also have an idea to design an powerful generalizing approach to various kinds of road networks with the potential of applying it to a bunch of cities.      

\section*{Acknowledgements}
The work was supported by the Joint Stock Company "Sberbank of Russia". 

%
%

\bibliographystyle{splncs03} 
\bibliography{8-refs}

\end{document}